\documentclass[10pt,twocolumn,letterpaper]{article}

\usepackage{cvpr}              

\usepackage{graphicx}
\usepackage{amsmath}
\usepackage{amssymb}
\usepackage{booktabs}
\usepackage[table]{xcolor}

\usepackage[pagebackref,breaklinks,colorlinks]{hyperref}

\usepackage[capitalize]{cleveref}
\crefname{section}{Sec.}{Secs.}
\Crefname{section}{Section}{Sections}
\Crefname{table}{Table}{Tables}
\crefname{table}{Tab.}{Tabs.}

\def\cvprPaperID{4450} 
\def\confName{CVPR}
\def\confYear{2022}

\begin{document}

\title{Temporal Feature Alignment and Mutual Information Maximization for \\ Video-Based Human Pose Estimation}

\author{
    Zhenguang Liu\textsuperscript{\rm 1},
    Runyang Feng\textsuperscript{\rm 2}\thanks{Corresponding Authors},
    Haoming Chen\textsuperscript{\rm 2*},
    Shuang Wu\textsuperscript{\rm 3*},
    Yixing Gao\textsuperscript{\rm 4},
    Yunjun Gao\textsuperscript{\rm 1},
    Xiang Wang\textsuperscript{\rm 5}
    \\    
    \textsuperscript{1} Zhejiang University, 
    \textsuperscript{2} Zhejiang Gongshang University, 
    \textsuperscript{3} Black Sesame Technologies,\\
    \textsuperscript{4} Jilin University, 
    \textsuperscript{5} National University of Singapore
	\\
	{\tt\small \{liuzhenguang2008, runyang2019.feng, chenhaomingbob\}@gmail.com,
	 \tt\small wushuang@outlook.sg,}
	 \\
	 {\tt\small gaoyixing@jlu.edu.cn, gaoyj@zju.edu.cn, xiangwang1223@gmail.com
	}
}
\maketitle

\begin{abstract}
Multi-frame human pose estimation has long been a compelling and fundamental problem in computer vision. This task is challenging due to fast motion and pose occlusion that frequently occur in videos. State-of-the-art methods strive to incorporate additional visual evidences from neighboring frames (supporting frames) to facilitate the pose estimation of the current frame (key frame). One aspect that has been obviated so far, is the fact that current methods directly aggregate unaligned contexts across frames. The spatial-misalignment between pose features of the current frame and neighboring frames might lead to unsatisfactory results. More importantly, existing approaches build upon the straightforward pose estimation loss, which unfortunately cannot constrain the network to fully leverage useful information from neighboring frames. 

To tackle these problems, we present a novel hierarchical alignment framework, which leverages coarse-to-fine deformations to progressively update a neighboring frame to align with the current frame at the feature level. We further propose to explicitly supervise the knowledge extraction from neighboring frames, guaranteeing that useful complementary cues are extracted. To achieve this goal, we theoretically analyzed the mutual information between the frames and arrived at a loss that maximizes the task-relevant mutual information. These allow us to rank No.1 in the Multi-frame Person Pose Estimation Challenge on benchmark dataset PoseTrack2017, and obtain state-of-the-art performance on benchmarks Sub-JHMDB and PoseTrack2018. Our code 
is released at \url{https://github.com/Pose-Group/FAMI-Pose}, hoping that it will be useful to the community.
\end{abstract}
\section{Introduction}
A key component of our capacity to interact with others lies in our ability to recognize the poses of humans \cite{schmidtke2021unsupervised, liu2021aggregated,iccv_motiontrajectory}. 
Likewise, detecting human poses is crucial for an intelligent machine to adjust its action and properly allocate its attention when interacting with people. Nowadays, pose estimation finds abundant applications in a wide spectrum of scenarios including action recognition, augmented reality, surveillance, and tracking \cite{luo2018lstm, yang2021learning}.
\begin{figure}[t]
\begin{center}
\includegraphics[width=0.98\linewidth]{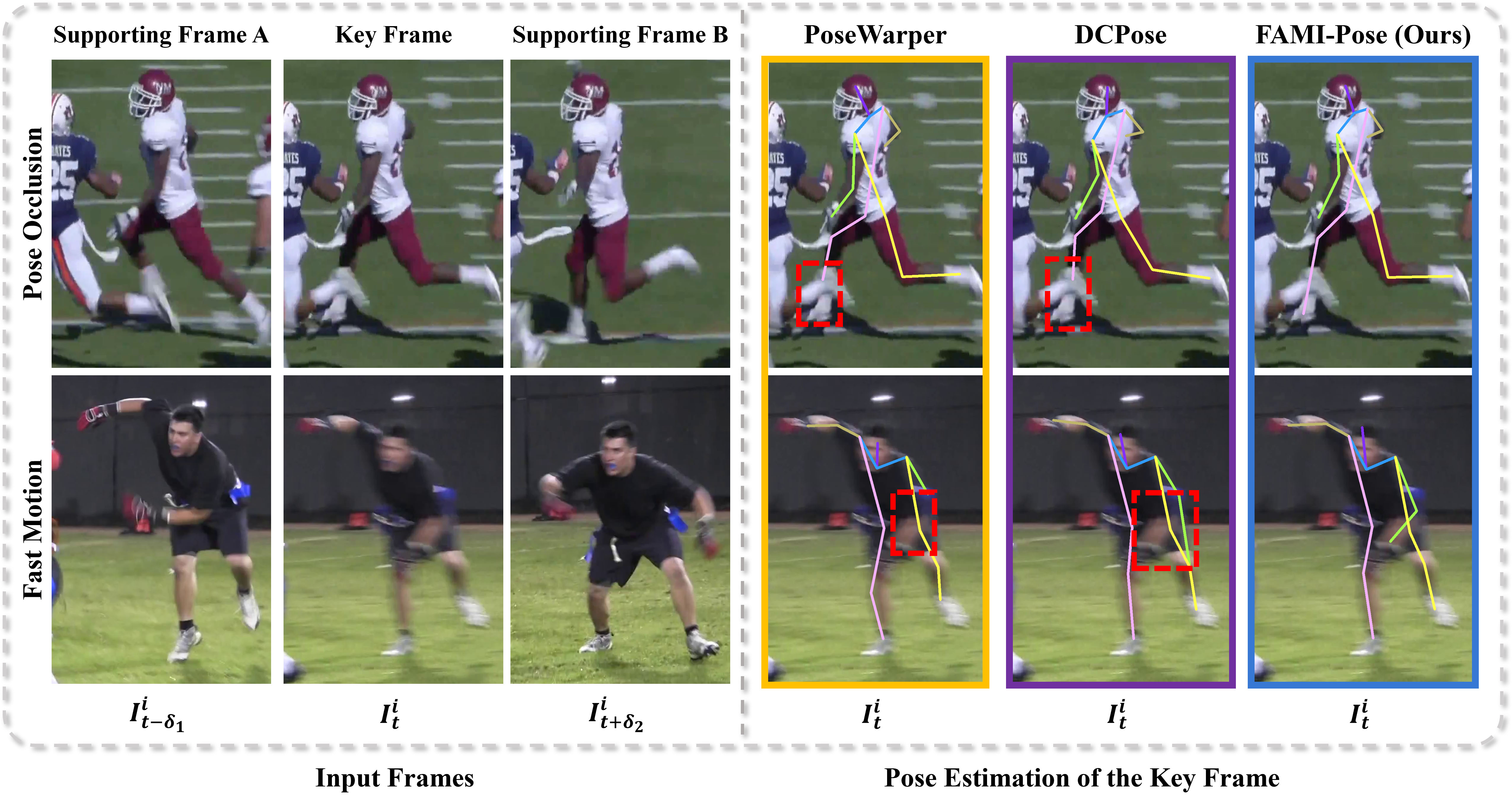}
\end{center}
\caption{State-of-the-art methods like PoseWarper and DCPose directly aggregate unaligned contexts from neighboring frames, which may fail for scenes with fast motion or pose occlusion. We perform temporal feature alignment between each supporting frame and the key frame, delivering robust pose estimations.} 
\label{fig:align}
\end{figure}

An extensive body of literature focuses on pose estimation in \emph{static images}, ranging from earlier approaches \cite{wang2008multiple, wang2013beyond, zhang2009efficient, sapp2010cascaded} utilising tree models or random forest models to recent attempts employing deep convolutional neural networks \cite{Cao_2017_CVPR, Toshev_2014_CVPR, Wei_2016_CVPR, newell2016stacked}. For pose estimation in videos, such methods are severely challenged in handling deteriorated video frames arising from scenes with fast motion and pose occlusion. Incorporating and leveraging additional contexts from neighboring frames is desirable to fill in the absent motion dynamics within a single frame and facilitate pose estimation. 

One line of work \cite{wang2020combining, luo2018lstm, artacho2020unipose} proposes to aggregate \emph{vanilla} sequential features of neighboring frames (supporting frames). \cite{luo2018lstm} trains a convolutional LSTM to model both spatial and temporal features, and directly predicts pose sequences for videos. \cite{wang2020combining} presents a 3D-HRNet to assemble features over a tracklet. Another line of work \cite{song2017thin, pfister2015flowing, liu2021deep} employs optical flow or implicit motion estimation to polish the pose estimation of the current frame (key frame). \cite{song2017thin, pfister2015flowing} propose to compute dense optical flow between frames, and leverage the flow based motion field for refining pose heatmaps temporally across multiple frames. \cite{liu2021deep} aggregates the pose heatmaps of consecutive frames and models motion residuals to improve pose estimation of the key frame.

Upon scrutinizing and experimenting on the released implementations of existing methods \cite{liu2021deep, bertasius2019learning, Dosovitskiy_2015_ICCV}, we observe that they suffer from performance deterioration in challenging cases such as rapid motion and pose occlusion. As illustrated in Fig. \ref{fig:align}, in the pose occlusion scenario, existing methods like DCPose fail to recognize the right ankle of the occluded person, leading to unexpected results. In the fast motion scenario, existing methods encounter difficulties in identifying the left wrist due to motion blur. 
We conjecture that the reasons are \textbf{twofolds}. \textbf{(1)} It is common that the same person in the current frame and a neighboring frame is not well aligned, especially for situations involving rapid motion of human subjects or cameras. However, existing methods tend to directly aggregate unaligned contexts from neighboring frames, these spatially misaligned features potentially diminish the performances of models.  
\textbf{(2)} State-of-the-art approaches simply employ the conventional MSE (Mean Square Error of joints) loss to supervise the learning of pose heatmaps, while lacking an effective constraint on guaranteeing information gain from neighboring frames as well as a supervision at the intermediate feature level.  

In this paper, we present a novel framework, along with theoretical analysis, to tackle the above challenges. The proposed method, 
termed FAMI-Pose (\textbf{\underline{F}}eature \textbf{\underline{A}}lignment and \textbf{\underline{M}}utual \textbf{\underline{I}}nformation maximization for \textbf{\underline{P}}ose estimation), consists of two key components. \textbf{(i)} FAMI-Pose conducts coarse-to-fine deformations that systematically update a neighboring frame to align with the current frame at the feature level. Specifically, FAMI-Pose first performs a \emph{global transformation}, which holistically rearranges neighboring frame feature to preliminarily rectify spatial shifts or jitter. Subsequently, a \emph{local calibration} is exploited to adaptively move and modulate each pixel of neighboring frame feature for enhanced feature alignment. \textbf{(ii)} FAMI-Pose further engages an information-theoretic objective as an additional intermediate supervision at the feature level. Maximizing this mutual information objective allows our model to fully mine task-relevant cues within the neighboring frames, extracting purposeful complementary knowledge to enhance pose estimation on the key frame. To the best of our knowledge, we are the first to methodically investigate the problem of feature alignment in human pose estimation and provide  insights from an  information-theoretic perspective.

We extensively evaluate the proposed method on three widely used benchmark datasets, PoseTack2017, PoseTrack2018, and Sub-JHMDB. Empirical evaluations show that our approach significantly outperforms current state-of-the-art methods. 
Our method achieves \textbf{84.8} mAP, \textbf{82.2} mAP, and \textbf{96.0} mAP on PoseTrack2017, PoseTrack2018, and Sub-JHMDB, respectively. 
Our results are submitted to the official evaluation server of PoseTack2017, and rank \emph{No.1} for this large benchmark dataset. 
We also present extensive ablation analyses on the contribution of each component, and validate the efficacy of feature alignment and the proposed mutual information loss. 

The contributions of this work are summarized as:
\begin{itemize}
	\item We propose to examine the multi-frame human pose estimation task from the perspective of effectively leveraging temporal contexts through feature alignment. 
	\item To explicitly supervise the knowledge extraction from neighboring frames, we propose an information-theoretic loss function, which allows maximizing the task-relevant cues mined from supporting frames.
	\item Our approach sets new state-of-the-art results on three benchmark datasets, PoseTrack2017, PoseTrack2018, and Sub-JHMDB. Our source code has been released.
\end{itemize}

\section{Related Work}
In this section, we briefly review the following three topics that are closely related to our work, namely image-based human pose estimation, video-based human pose estimation, and feature alignment.

\subsection{Image-Based Human Pose Estimation}
Conventional solutions to image-based human pose estimation utilize pictorial structures  \cite{zhang2009efficient, sapp2010cascaded} to model the spatial relationships among body joints. These methods tend to rely on handcrafted features and have limited representational abilities.
Fueled by the explosion of deep learning \cite{wang2020combining,hao2020person} and the availability of large-scale pose estimation datasets such as PoseTrack \cite{Iqbal_2017_CVPR, Andriluka_2018_CVPR} and COCO \cite{lin2014microsoft}, various deep learning methods 
\cite{artacho2020unipose, cheng2020higherhrnet, huang2020devil, zhang2020distribution, varamesh2020mixture, su2019multi, vqa1, vqa2, yang2017person, yang2021deconfounded} have been proposed. These methods can be broadly categorized into two paradigms: bottom-up and top-down. \emph{Bottom-up approaches} \cite{Cao_2017_CVPR, kocabas2018multiposenet, kreiss2019pifpaf,li2019crowdpose} first detect individual body parts, and then assemble these detected constituent parts into the entire person. \cite{Cao_2017_CVPR} proposes a dual convolution structure to simultaneously predict part confidence maps and part affinity fields (that represent the relationships between body parts). On the other hand, \emph{top-down approaches} \cite{xiao2018simple, Wei_2016_CVPR, sun2019deep, newell2016stacked, moon2019posefix} first detect human bounding boxes and then estimate human poses within each bounding box. \cite{xiao2018simple} leverages deconvolution layers to replace the commonly used bi-linear interpolation for spatial-upsampling of feature maps. A recent work in \cite{sun2019deep} presents a high resolution network (HRNet) that retains high resolution feature maps throughout the entire inference, achieving state-of-the-art results on multiple image-based benchmarks.

\subsection{Video-Based Human Pose Estimation}
Pose estimation models trained for image-based data could not generalize well to video sequences due to their inability to incorporate abundant cues from neighboring frames. To model and leverage temporal contexts across frames, one direct approach would be employing convolutional LSTMs as proposed in \cite{luo2018lstm, artacho2020unipose}. A key shortcoming of such models might be their tendency to misalign features across different frames, which unfavourably   reduces the potency of the supporting frames. \cite{song2017thin,  pfister2015flowing} explicitly estimate motion fields by computing optical flow between consecutive frames, and these motion cues are subsequently used for aligning pose heatmaps. \cite{liu2021deep} estimates motion offsets between the key frame and supporting frames, and these offsets provide the basis to perform resampling of pose heatmaps on consecutive frames. In both cases, the pose estimation accuracy would be heavily dependent on the performance of the optical flow or motion offset estimation. Furthermore, the lack of an effective supervision at the intermediate features level for these approaches could lead to inaccurate pose estimations.

\begin{figure*}
\begin{center}
\includegraphics[width=1\linewidth]{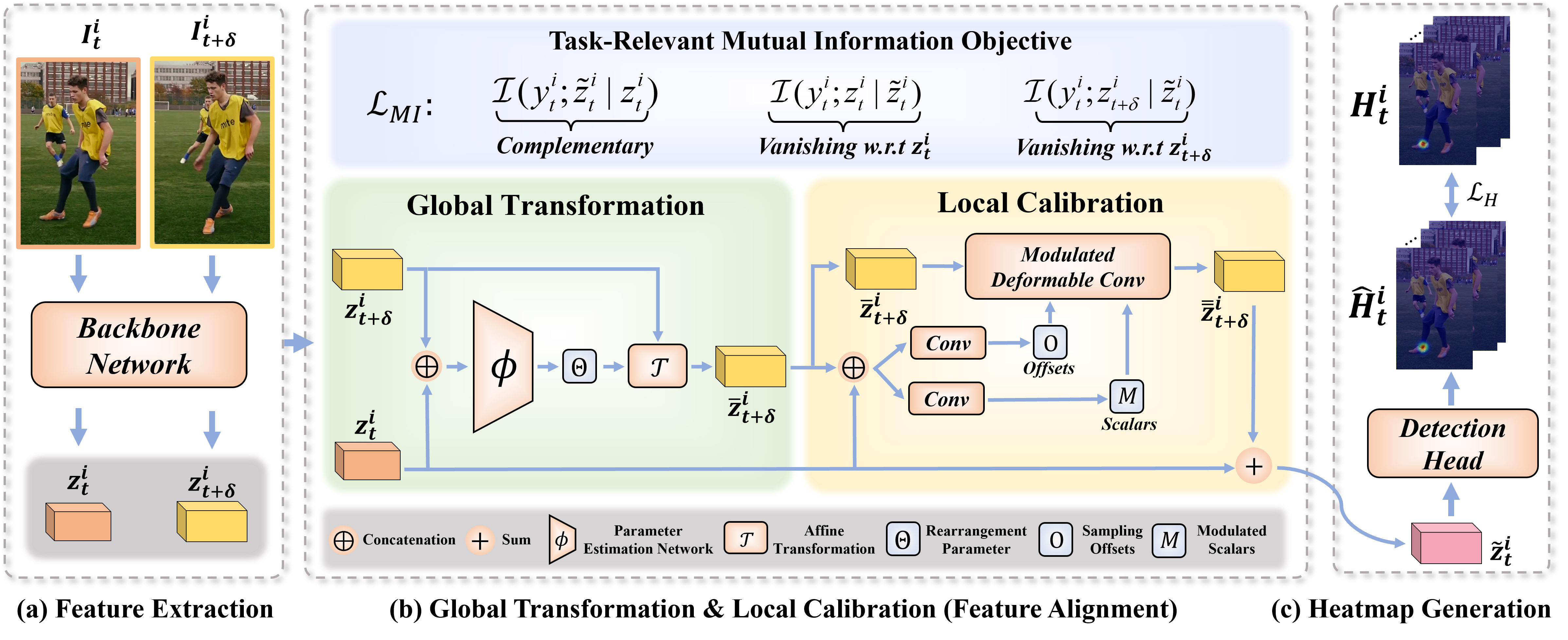}
\end{center}
\caption{Overall pipeline of our FAMI-Pose framework. The goal is to detect the pose of person $i$ in the key frame $I_t^i$, with the assistance of its supporting frames.  For clarity of illustration, we only show a single supporting frame $I_{t+\delta}^i$ in this figure. We first extract their respective features $z_t^i$ and $z_{t+\delta}^i$. These features are then handed to our global transformation module and the local calibration module for temporal alignment. The key frame feature $z^i_t$ and aligned features $\bar{\bar{z}}^i_{t+\delta}$ for all supporting frames are aggregated to $\tilde{z}^i_t$, which is passed to a detection head that outputs pose estimates $\widehat{H}^i_t$. Besides the heatmap estimation loss $\mathcal{L}_{H}$ that measures the discrepancy between $\widehat{H}^i_t$ and the ground truth $H^i_t$, we introduce an additional feature level supervision through our Mutual Information objective $\mathcal{L}_{MI}$ to extract maximal task-relevant complementary information from supporting frames.}
\label{fig:pipeline}
\end{figure*}

\subsection{Feature Alignment}
Feature alignment is an important topic for many computer vision tasks (\emph{e.g.}, semantic segmentation \cite{mazzini2018guided, li2020semantic}, object detection \cite{he2017mask, chen2019revisiting}), and numerous efforts have recently been made to address this problem. \cite{lu2019indices} presents an index-guided framework that employs indices to guide the pooling and upsampling. \cite{huang2021fapn} proposes to learn the transformation offsets of pixels to align upsampled feature maps. \cite{huang2021alignseg} presents an aligned feature aggregation module to align the features of multiple different resolutions for better aggregation. Whereas previous methods mostly tackle spatial misalignment between network inputs and outputs, we focus on temporal (\emph{i.e.}, across frames) feature alignment in this work. 

\section{Our Approach}

\textbf{Preliminaries}\quad 
To detect human poses from the video frames, we first extract the bounding box of each individual person. Technically, for a video frame $I_t$, we first employ an object detector to extract the bounding box for each individual person. This bounding box is then enlarged by 25\% to crop the same individual on a predefined window $\mathcal{N}$ of neighboring frames. Overall, for person $i$, we obtain the cropped image $I_t^i$ for the key frame and $\{I_{t+\delta}^i \mid \delta\in\mathcal{N}\}$ for the supporting (neighboring) frames.

\textbf{Problem Formulation}\quad
Presented with a key frame $I_t^i$ along with its supporting frames $\{I_{t+\delta}^i\mid\delta\in \mathcal{N}\}$, our goal is to estimate the pose in $I_t^i$. We seek to better leverage the supporting frames through a principled feature alignment and mining task relevant information, thereby addressing the common drawback of existing approaches in failing to adequately tap into the temporal information.


\textbf{Method Overview}\quad
An overview of our pipeline is illustrated in Fig. \ref{fig:pipeline}. For each supporting frame $I_{t+\delta}^i$, FAMI-pose performs a two-stage hierarchical transformation to align $I_{t+\delta}^i$ with the key frame $I_t^i$ at the feature level. Specifically, FAMI-Pose consists of two main modules, a global transformation module and a local calibration module. We first perform feature extraction on ${I}^{i}_{t}$ and ${I}^{i}_{t+\delta}$ to obtain ${z}^{i}_{t}$ and ${z}^{i}_{t+\delta}$, respectively. These features are then fed into our global transformation module, which learns the parameters of an affine transformation to obtain a coarsely aligned supporting frame feature $\bar{z}^{i}_{t+\delta}$. ${z}^{i}_{t}$ and $\bar{z}^{i}_{t+\delta}$ are then handed to the local calibration module, which performs pixel-wise deformation to produce finely aligned features $\bar{\bar{z}}^{i}_{t+\delta}$. Finally, we aggregate all aligned supporting frame features $\{\bar{\bar{z}}^{i}_{t+\delta}\mid\delta\in\mathcal{N}\}$ and the key frame feature $z_t^i$ to obtain our enhanced feature $\tilde{z}^{i}_{t}$. $\tilde{z}^{i}_{t}$ is passed to a detection head that outputs pose estimations $\widehat{H}^i_t$. The task objective is to minimize the heatmap estimation loss $\mathcal{L}_H$ which measures the discrepancy between $\widehat{H}^i_t$ and the ground truth ${H}^i_t$. On top of this, we also design a mutual information objective $\mathcal{L}_{MI}$ which effectuates a feature level supervision for maximizing the amount of complementary task-relevant information encoded in $\tilde{z}^{i}_{t}$. In what follows, we introduce the complete FAMI-Pose architecture and the mutual information objective in detail.

\subsection {Feature Alignment}
Feature alignment starts with feature extraction, which  is done with the HRNet-W48 network \cite{sun2019deep} (the state-of-the-art method for image-based human pose estimation) as the backbone. The extracted features ${z}^{i}_{t}$ and ${z}^{i}_{t+\delta}$ are then passed through a global transformation module and a local calibration module, to progressively align ${z}^{i}_{t+\delta}$ with ${z}^{i}_{t}$. We would like to highlight that we do not pursue an  image-level alignment, instead we drive the network to learn a feature-level alignment between a supporting frame and the key frame.

\textbf{Global Transformation}\quad
We observe that most failure cases for pose estimation in videos occur due to rapid movements of persons or cameras, which inevitably lead to large spatial shifts or jitters between neighboring frames. In order to align a supporting frame to the key frame, we design a global transformation module (GTM). The GTM computes spatial rearrangement parameters of a global affine transformation to obtain a coarse preliminary alignment of supporting frame feature ${z}^{i}_{t+\delta}$ with the key frame feature ${z}^{i}_{t}$.

More specifically, the GTM includes two submodules: 
\begin{enumerate}
\item A spatial rearrangement parameter estimation network $\phi$ that estimates affine transformation parameters $\Theta$ from the input feature pair as $\phi:({z}^{i}_{t}, {z}^{i}_{t+\delta})  \rightarrow \Theta \in \mathbb{R}^{2\times3}$. The elements of $\Theta$ correspond to translation, rotation, shear, and scaling operations.
\item Subsequently, a global affine transformation $\mathcal{T}$ is performed to obtain the preliminarily aligned supporting frame feature $\mathcal{T}: ({z}^{i}_{t+\delta}, \Theta) \rightarrow \bar{{z}}^{i}_{t+\delta}$.
\end{enumerate}

The operations of the GTM can be expressed as follows:
\begin{equation}
\begin{aligned}
\Theta
&=\phi \left( {z}_{t}^i \oplus {z}_{t+\delta}^i \right),\\
\left(\begin{array}{l}
{x}_{p} \\
{y}_{p} \\
\end{array}\right)&=\underbrace{\left[
\begin{array}{lll}
\theta_{11} & \theta_{12} & \theta_{13} \\
\theta_{21} & \theta_{22} & \theta_{23}
\end{array}
\right]}_{\Theta}
\left(
\begin{array}{c}
\bar{x}_{p} \\
\bar{y}_{p} \\
1
\end{array}\right),
\end{aligned}
\end{equation}
where $(x_{p}, y_{p})$ and $(\bar{x}_{p}, \bar{y}_{p})$ denote the coordinates of pixel $p$ for ${z}_{t+\delta}^i$ and $\bar{z}_{t+\delta}^i$, respectively.



\textbf{Local Calibration}\quad
The global transformation module produces a coarse alignment. We then design our local calibration module (LCM) to perform meticulous fine-tuning at a pixel-level, yielding finely aligned features $\bar{\bar{z}}^{i}_{t+\delta}$.

Specifically, given $\bar{z}^{i}_{t+\delta}$ and ${z}^{i}_{t}$, we independently estimate convolution kernel sampling offsets $O$ and modulated scalars $M$ for the feature $\bar{z}^{i}_{t+\delta}$: 
\begin{equation}
\begin{aligned}
\bar{z}^{i}_{t+\delta} \oplus {z}^{i}_{t}  &\xrightarrow[\text{blocks}]{\text{residual}} \xrightarrow[\text{convolution}]{\text{regular}} O,\\
\bar{z}^{i}_{t+\delta} \oplus {z}^{i}_{t}  &\xrightarrow[\text{blocks}]{\text{residual}} \xrightarrow[\text{convolution}]{\text{regular}} M.
\end{aligned}
\end{equation}
The adaptively learned kernel offsets $O$ and modulated scalars $M$ respectively correspond to \emph{location shifts} and \emph{intensity fluctuations} of each pixel in $\bar{z}^{i}_{t+\delta}$ with respect to the key frame feature ${z}^{i}_{t}$.

Subsequently, we implement the local calibration operation through the modulated deformable convolution \cite{zhu2019deformable}. Given the preliminarily aligned features $\bar{z}^{i}_{t+\delta}$, the kernel sampling offsets $O$, and the modulated scalars $M$ as inputs, the modulated deformable convolution outputs the fine-tuned feature $\bar{\bar{z}}^{i}_{t+\delta}$:
\begin{equation}
\begin{aligned}
\left(\bar{z}^{i}_{t+\delta}, O, M \right) 
\xrightarrow[\text{convolution}]{\text{modulated deformable}}
\bar{\bar{z}}^{i}_{t+\delta}.
\end{aligned}
\end{equation}

To anticipate the discussion of the mutual information loss, we would like to point out that the key frame feature $z^{i}_{t}$ is only used for computing the global transformation parameters in GTM and convolutional parameters in LCM. Its information will not be propagated into the final aligned supporting frame feature $\bar{\bar{z}}^{i}_{t+\delta}$.

\textbf{Heatmap Generation}\quad
Ultimately, we aggregate over all final aligned supporting frame features $\{\bar{\bar{z}}^{i}_{t+\delta}\mid\delta\in\mathcal{N}\}$ and the key frame feature ${z}^{i}_{t}$ via element-wise addition to obtain the enhanced feature $\tilde{z}^{i}_{t}$. $\tilde{z}^{i}_{t}$ is fed to a detection head to produce pose heatmap estimations $\widehat{H}_t^{i}$. We implemented the detection head using a stack of $3\times 3$ convolutions. By effectively leveraging temporal information from supporting frames through our coarse-to-fine alignment modules, our FAMI-Pose is more adept at tackling visual degeneration issues and therefore gives more accurate pose heatmaps.

\subsection {Mutual Information Objective}
\label{MI}
We can certainly train the FAMI-Pose in a direct end-to-end manner with a pose heatmap loss, as is done in most previous methods \cite{sun2019deep, xiao2018simple, liu2021deep, wang2020combining, bertasius2019learning}. Given our systematic examination of extracting temporal features for pose estimation, it would be fruitful to investigate whether introducing \emph{supervision at the feature level} would facilitate the task.

Naively, we could formulate the feature level objective as the $L1$ or $L2$ difference between supporting frames feature $z_{t+\delta}^i$ and the key frame feature $z_{t}^i$. However, such rigid-alignment is likely to lead to erosion of complementary task-specific information from supporting frames. Consequently, the temporal features thus optimized would be inadequate for providing relevant supporting information to facilitate pose estimation. 

It is therefore crucial that we highlight the purposeful complementary information from the supporting frames. Towards this end, inspired by \cite{zhao2021learning, hjelm2018learning}, we propose a mutual information objective, which seeks to maximize the amount of complementary task-relevant information in the enhanced feature $\tilde{z}_t^i$. 

\textbf{Mutual Information}\quad
Mutual information (MI) is a measure of the amount of information shared between random variables.
Formally, MI quantifies the statistical dependency of two random variables $\boldsymbol{v}_{1}$ and $\boldsymbol{v}_{2}$:
\begin{equation}
\mathcal{I}(\boldsymbol{v}_{1} ; \boldsymbol{v}_{2})
=
\mathbb{E}_{p(\boldsymbol{v}_{1}, \boldsymbol{v}_{2})}\left[\log \frac{p(\boldsymbol{v}_{1}, \boldsymbol{v}_{2})}{p(\boldsymbol{v}_{1}) p(\boldsymbol{v}_{2})}\right],
\end{equation}
where $p(\boldsymbol{v}_{1}, \boldsymbol{v}_{2})$ is the joint probability distribution between $\boldsymbol{v}_{1}$ and $\boldsymbol{v}_{2}$,
while $p(\boldsymbol{v}_{1})$ and $p(\boldsymbol{v}_{2})$ are their marginals.

\textbf{Mutual Information Loss} \quad
Within this framework, our primary objective for learning effective temporal feature alignment can be formulated as:
\begin{equation}
\begin{aligned}
\label{eq.max}
\text{max } \mathcal{I}\left({y}^{i}_{t} ; \tilde{z}^{i}_{t} \mid {z}^{i}_{t} \right),
\end{aligned}
\end{equation}
where ${y}^{i}_{t}$ represents the label, and $\mathcal{I}\left({y}^{i}_{t} ; \tilde{z}^{i}_{t} \mid {z}^{i}_{t} \right) $ denotes the amount of task-relevant information in the enhanced feature $\tilde{z}^{i}_{t}$, complementary to (\emph{i.e.}, excluding) the information from the key frame feature ${z}^{i}_{t}$. Intuitively, optimizing this objective will maximize the additional relevant and complementary information we seek to extract from neighboring frames to support the pose estimation task.

Due to the notorious difficulty of the conditional MI computations especially in neural networks \cite{hjelm2018learning, tian2021farewell}, we perform a simplification. We first factorize Eq. \ref{eq.max} as follows:
\begin{equation}
\begin{aligned}
\label{eq.com.}
\mathcal{I}     \left({y}^{i}_{t}                   ; \tilde{z}^{i}_{t}  \mid    {z}^{i}_{t} \right)  
= \mathcal{I} \left( {y}^{i}_{t}                   ; \tilde{z}^{i}_{t}  \right)
&-  \mathcal{I} \left( \tilde{z}^{i}_{t} ; {z}^{i}_{t}  \right)\\
&+ \mathcal{I} \left( \tilde{z}^{i}_{t} ; {z}^{i}_{t}                    \mid   {y}^{i}_{t}  \right),
\end{aligned} 
\end{equation}
where $\mathcal{I} \left( {y}^{i}_{t};\tilde{z}^{i}_{t}  \right)$ measures the relevance of the label ${y}^{i}_{t}$ and feature $\tilde{z}^{i}_{t}$, $\mathcal{I} \left( \tilde{z}^{i}_{t} ; {z}^{i}_{t}  \right)$ indicates the dependence between the two features $\tilde{z}^{i}_{t}$ and ${z}^{i}_{t}$, and $\mathcal{I} \left( \tilde{z}^{i}_{t} ; {z}^{i}_{t} \mid {y}^{i}_{t} \right)$ represents the \emph{task-irrelevant information} in both $\tilde{z}^{i}_{t}$ and ${z}^{i}_{t}$. Heuristically, when optimizing over the task objective, the task-specific information will have an overwhelming presence over the task-irrelevant information. Therefore, we may assume that the task-irrelevant information will be negligible upon sufficient training \cite{zhao2021learning, federici2019learning}. This simplifies Eq. \ref{eq.com.} to:
\begin{equation}
\begin{aligned}
\label{eq:Simplification of MI}
\mathcal{I}\left(  {y}^{i}_{t} ; \tilde{z}^{i}_{t}  \mid    {z}^{i}_{t} \right)  
\rightarrow
\mathcal{I} \left( {y}^{i}_{t} ;  \tilde{z}^{i}_{t} \right)
-  
\mathcal{I} \left( {z}^{i}_{t} ;  \tilde{z}^{i}_{t}  \right).
\end{aligned}
\end{equation}
Moreover, we introduce two regularization terms to alleviate information dropping: 
\begin{equation}
\begin{aligned}
\label{eq.re.}
\min \left[ 
\mathcal{I}\left({y}^{i}_{t} ;  {z}^{i}_{t+\delta}  \mid    \tilde{z}^{i}_{t} \right)
+ 
\mathcal{I}\left({y}^{i}_{t} ;  {z}^{i}_{t}             \mid    \tilde{z}^{i}_{t} \right)
\right].
\end{aligned} 
\end{equation}
The terms $\mathcal{I}\left({y}^{i}_{t} ;  {z}^{i}_{t+\delta}  \mid \tilde{z}^{i}_{t} \right)$ and $\mathcal{I}\left({y}^{i}_{t} ;  {z}^{i}_{t}             \mid    \tilde{z}^{i}_{t} \right)$ respectively measure the vanishing task-relevant information in ${z}^{i}_{t+\delta}$ and $z^i_t$ during feature alignment. They serve to facilitate the nondestructive propagation of information. Simultaneously minimizing these two terms would prevent excessive information loss in ${z}^{i}_{t+\delta}$ and $z^i_t$ while maximizing the primary complementary task-relevant mutual information objective. 

Similar to Eq. \ref{eq:Simplification of MI}, we simplify the two regularization terms in Eq. \ref{eq.re.} as follows: 
\begin{equation}
\begin{aligned}
\mathcal{I} \left({y}^{i}_{t}           ;  {z}^{i}_{t+\delta}  \mid \tilde{z}^{i}_{t} \right) 
&\rightarrow
\mathcal{I} \left({y}^{i}_{t}           ;  {z}^{i}_{t+\delta}  \right)
-  
\mathcal{I} \left({z}^{i}_{t+\delta} ;  \tilde{z}^{i}_{t}  \right),
\\
\mathcal{I} \left({y}^{i}_{t}            ; {z}^{i}_{t}  \mid \tilde{z}_{t} \right)
&\rightarrow
\mathcal{I} \left( {y}^{i}_{t}          ; {z}^{i}_{t}  \right)
-  
\mathcal{I} \left( {z}^{i}_{t}          ;  \tilde{z}^{i}_{t}  \right).
\end{aligned} 
\end{equation}

Finally, we simultaneously optimize the complementary information term in Eq. \ref{eq.max} and the two regularization terms in Eq. \ref{eq.re.} to provide feature level supervision:
\begin{equation}
\begin{aligned}
\mathcal{L}_\text{MI} =
\overbrace{\mathcal{I}\left({y}^{i}_{t}   ; {z}^{i}_{t}                       \mid   \tilde{z}^{i}_{t}  \right)}^{\text{Vanishing w.r.t. }{z}^{i}_{t}}
&+  
\overbrace{\mathcal{I}\left({y}^{i}_{t}   ;  {z}^{i}_{t+\delta}            \mid   \tilde{z}^{i}_{t}  \right)}^{\text{Vanishing w.r.t. }{z}^{i}_{t+\delta}}
\\
&-  \alpha \cdot 
\underbrace{\mathcal{I}\left({y}^{i}_{t}   ;   \tilde{z}^{i}_{t}  \mid   {z}^{i}_{t} \right)}_\text{Complementary},
\end{aligned} 
\label{alpha}
\end{equation}
where $\alpha$ serves as a hyper-parameter in our network to balance the ratios of different terms. These MI terms can be estimated by existing MI estimators \cite{belghazi2018mutual, tian2021farewell, van2018representation, cheng2020club}. In our experiments, we employ the Variational Self-Distillation (VSD) \cite{tian2021farewell} to estimate the MI for each term.

\subsection{Training Objective}
Our training objective consists of two parts. (1) We adopt the heatmap estimation loss function $\mathcal{L}_\text{H}$ to supervise the learning of final pose estimates:
\begin{equation}
\begin{aligned}
\mathcal{L}_\text{H}=\left\|
\widehat{H}_t^{i} - {H}_t^{i}\right\|_{2}^{2},
\end{aligned}
\end{equation}
where $\widehat{H}_t^{i}$ and ${H}_t^{i}$ denotes the prediction heatmap and ground truth heatmap, respectively.
(2) We also leverage the proposed MI loss to supervise the temporal features as described in Sec. \ref{MI}. The overall loss function is given by:
\begin{equation}
\begin{aligned}
\mathcal{L}_{total} =  \mathcal{L}_\text{H} + \beta \cdot \mathcal{L}_\text{MI}.
\end{aligned} 
\label{beta}
\end{equation}


\section{Experiments}
In this section, we present our experimental results on three widely used benchmark datasets, namely PoseTrack2017 \cite{Iqbal_2017_CVPR}, PoseTrack2018 \cite{Andriluka_2018_CVPR}, and Sub-JHMDB \cite{Jhuang:ICCV:2013}.

\renewcommand\arraystretch{1.2}
\begin{table}
  \resizebox{0.48\textwidth}{!}{
  \begin{tabular}{l|c|c|c|c|c|c|c|c}
    \hline
      Method                            &Head   &Shoulder &Elbow       &Wrist   &Hip    &Knee   &Ankle   &{\bf Mean}\cr
      \hline
      PoseTracker \cite{girdhar2018detect}   &$67.5$ &$70.2$   &$62.0$      &$51.7$  &$60.7$ &$58.7$ &$49.8$  &{$60.6$}\cr
     PoseFlow \cite{xiu2018pose}         &$66.7$ & $73.3$  &$68.3$      &$61.1$  &$67.5$ &$67.0$ &$61.3$  &{$ 66.5$}\cr
JointFlow \cite{doering2018joint}        & -     & -       &-           &-       &-      &-      &-       &{ $ 69.3$}\cr
   FastPose \cite{zhang2019fastpose}   	&$80.0$ &$80.3$   &$69.5$      &$59.1$  &$71.4$ &$67.5$ &$59.4$  &{$ 70.3$}\cr
   TML++ \cite{hwang2019pose}    	 		&-       &-     &-      &-      &-      &-       &-    &{$ 71.5$}\cr
Simple (ResNet-50) \cite{xiao2018simple}    &$79.1$ &$80.5$   &$75.5$      &$66.0$  &$70.8$ &$70.0$ &$61.7$  &{$72.4$}\cr
Simple (ResNet-152) \cite{xiao2018simple}    &$81.7$ &$83.4$   &$80.0$      &$72.4$  &$75.3$ &$74.8$ &$67.1$  &{$ 76.7$}\cr
  STEmbedding \cite{jin2019multi}        &$83.8$ &$81.6$   &$77.1$      &$70.0$  &$77.4$ &$74.5$ &$70.8$  &{$ 77.0$}\cr
        HRNet \cite{sun2019deep}         &$82.1$ &$83.6$   &$80.4$      &$73.3$  &$75.5$ &$75.3$ &$68.5$  &{$ 77.3$}\cr
         MDPN \cite{guo2018multi}        &$85.2$ &$88.5$   &$83.9$      &$77.5$  & $79.0$&$77.0$ &$71.4$  &{$ 80.7$}\cr
   Dynamic-GNN \cite{yang2021learning} 	 &$88.4$ &$88.4$   &$82.0$      &$ 74.5$ &$79.1$ &$78.3$ &$73.1$  &{ $81.1$}\cr
 PoseWarper \cite{bertasius2019learning} &$81.4$ &$88.3$   &$83.9$      &$ 78.0$ &$82.4$ &$80.5$ &$73.6$  &{ $ 81.2$}\cr
   DCPose \cite{liu2021deep}  &$ 88.0$  &$ 88.7$     &$ 84.1$   &$78.4$&$ 83.0$        &$ 81.4$&$ 74.2$ &$ 82.8$\cr
	\hline
    \rowcolor{gray!20} \bf FAMI-Pose (Ours)	&$\bf 89.6$  &$\bf 90.1$ &$\bf 86.3$ &$\bf 80.0$ &$\bf 84.6$ &$\bf 83.4$ &$\bf 77.0$ &$\bf 84.8$\cr 
    \hline
    \end{tabular}}
    \caption{{Quantitative results on the PoseTrack2017 \textbf{validation} set}.} \label{17val}
\end{table}
\renewcommand\arraystretch{1.2}
\begin{table}
  \resizebox{0.48\textwidth}{!}{
  \begin{tabular}{l|c|c|c|c|c|c|c|c}
    \hline
     Method                            &Head&Shoulder &Elbow  &Wrist &Hip &Knee &Ankle &{\bf Total}\cr
    \hline
PoseTracker \cite{girdhar2018detect}    &-       &-     &-      &$51.5$ &-      &-       &$50.2$   &{$ 59.6$}\cr
PoseFlow \cite{xiu2018pose}             &$64.9$  &$67.5$&$65.0$ &$59.0$ &$62.5$ &$62.8$  &$57.9$   &{$ 63.0$}\cr
JointFlow \cite{doering2018joint}       &-       &-     &-      &$53.1$ &-      &-       &$50.4$   &{$ 63.4$}\cr
 TML++ \cite{hwang2019pose}    	 		&-       &-     &-      &$60.9$ &-      &-       &$56.0$   &{$ 67.8$}\cr
KeyTrack \cite{snower202015}            &-       &-     &-      &$71.9$ &-      &-       &$65.0$   &{$ 74.0$}\cr
DetTrack \cite{wang2020combining}       &-       &-     &-      &$69.8$ &-      &-       &$65.9$   &{$ 74.1$}\cr
Simple (ResNet-152) \cite{xiao2018simple}    &$80.1$ &$80.2$ &$76.9$ &$71.5$ &$72.5$ &$72.4$  &$65.7$   &{$ 74.6$}\cr
HRNet \cite{sun2019deep}                &$80.1$ &$80.2$ &$76.9$ &$72.0$ &$73.4$ &$72.5$  &$67.0$   &{$ 74.9$}\cr
PoseWarper \cite{bertasius2019learning} &$79.5$ &$84.3$ &$80.1$ &$75.8$ &$77.6$ &$76.8$  &$70.8$   &{$ 77.9$}\cr
   DCPose \cite{liu2021deep} &$84.3$	  &$ 84.9$   &$ 80.5$   &$ 76.1$   &$ 77.9$   &$ 77.1$   &$ 71.2$   &$ 79.2$\cr    \hline
    \rowcolor{gray!20}\bf FAMI-Pose (Ours)&$\bf 86.1$&$\bf 86.1$&$\bf 81.8$&$\bf 77.4$&$\bf 79.5$&$\bf 79.1$&$\bf 73.6$&$\bf 80.9$\cr
    \hline
    \end{tabular}}
    \caption{Performance comparisons on the PoseTrack2017 \textbf{test} set. These results are published in the \emph{PoseTrack2017 leaderboard}.} \label{17test}
\end{table}

\renewcommand\arraystretch{1.2}
\begin{table}
   \resizebox{0.48\textwidth}{!}{
   \begin{tabular}{l|c|c|c|c|c|c|c|c}
     \hline
      Method                            &Head &Shoulder &Elbow  &Wrist &Hip &Knee &Ankle &{\bf Mean}\cr
     \hline
  STAF \cite{raaj2019efficient}    	  	&-       &-     &-      &$64.7$ &-      &-       &$62.0$   &{$70.4$}\cr
 AlphaPose \cite{fang2017rmpe}           &$63.9$  &$78.7$&$77.4$ &$71.0$ &$73.7$ &$73.0$    &69.7     &{$71.9$}\cr
  TML++ \cite{hwang2019pose}    	 		&-       &-     &-      &-      &-      &-       &-    &{$ 74.6$}\cr
 MDPN \cite{guo2018multi}                &$75.4$ &$81.2$ &$79.0$ &$74.1$ &$72.4$ &$73.0$  &$69.9$   &{$75.0$}\cr
 PGPT \cite{bao2020pose}    	 		&-       &-     &-      &$72.3$ &-      &-       &$72.2$   &{$76.8$}\cr
 Dynamic-GNN \cite{yang2021learning} 	&$80.6$ &$84.5$   &$80.6$  &$ 74.4$ &$75.0$ &$76.7$ &$71.8$  &{ $77.9$}\cr
 PoseWarper \cite{bertasius2019learning} &$79.9$&$86.3$&$82.4$&$77.5$&$79.8$&$78.8$&$73.2$  &{ $79.7$}\cr
      DCPose \cite{liu2021deep}&$ 84.0$&$ 86.6$&$ 82.7$&$ 78.0$&$ 80.4$&$ 79.3$&$ 73.8$&$ 80.9$\cr \hline
     \rowcolor{gray!20}\bf FAMI-Pose (Ours)&$\bf 85.5$&$\bf 87.7$&$\bf 84.2$&$\bf 79.2$&$\bf 81.4$&$\bf 81.1$&$\bf 74.9$&$\bf 82.2$\cr
     \hline
     \end{tabular}}
     \caption{\footnotesize{Quantitative results on the PoseTrack2018 \textbf{validation} set}.} \label{18val}
   \end{table}

\renewcommand\arraystretch{1.2}
\begin{table}
  \resizebox{0.48\textwidth}{!}{
  \begin{tabular}{l|c|c|c|c|c|c|c|c}
    \hline
     Method                            &Head&Shoulder &Elbow  &Wrist &Hip &Knee &Ankle &{\bf Total}\cr
    \hline
     TML++ \cite{hwang2019pose}    	   &-       &-     &-      &$60.2$ &-      &-       &$56.9$   &{$ 67.8$}\cr
	 AlphaPose++ \cite{fang2017rmpe, guo2018multi}         &- &- &- &$66.2$ &- &-  &$65.0$     &{ $ 67.6$}\cr
	 DetTrack \cite{wang2020combining}      &-      &-      &-      &$69.8$ &-      &-       &$67.1$     &$73.5$\cr
 	MDPN \cite{guo2018multi}                &- &- &- &$74.5$ &- &-  &$69.0$   &{$ 76.4$}\cr
	PoseWarper \cite{bertasius2019learning} &$78.9$&$ 84.4$&$ 80.9$&$76.8$&$75.6$&$77.5$&$71.8$& $ 78.0$\cr
    DCPose \cite{liu2021deep} &$ 82.8$&$ 84.0$&$ 80.8$&$ 77.2$&$ 76.1$&$ 77.6$&$ 72.3$&$ 79.0$\cr
    \hline
    \rowcolor{gray!20}\bf FAMI-Pose (Ours)&$\bf 83.6$&$\bf 84.5$&$\bf 81.4$&$\bf 77.9$&$\bf 76.8$&$\bf 78.3$&$\bf 72.9$&$\bf 79.6$\cr
    \hline
    \end{tabular}}
    \caption{Performance comparisons on the PoseTrack2018 \textbf{test} set.} \label{18test}
\end{table}

\renewcommand\arraystretch{1.2}
\begin{table}
  \resizebox{0.48\textwidth}{!}{
  \begin{tabular}{l|c|c|c|c|c|c|c|c}
    \hline
     Method                            &Head&Shoulder &Elbow  &Wrist &Hip &Knee &Ankle &{\bf Avg}\cr
    \hline
Part Models\cite{park2011n}             &$79.0$  &$60.3$&$28.7$ &$16.0$ &$74.8$ &$59.2$  &$49.3$   &$52.5$\cr
Joint Action\cite{xiaohan2015joint}     &$83.3 $      &$63.5$     &$33.8$      &$21.6$ &$76.3$    &$62.7$      &$53.1$   &{$55.7$}\cr
Pose-Action\cite{iqbal2017pose}         &$90.3$       &$76.9$    &$59.3$      &$55.0$ &$85.9$   &$76.4$  &$73.0$   &{$73.8$}\cr
CPM\cite{wei2016convolutional}			&$98.4$       &$94.7$    &$85.5$      &$81.7$ &$97.9$   &$94.9$  &$90.3$   &{$91.9$}\cr
Thin-slicing Net\cite{song2017thin}     &$97.1$      &$95.7$     &$87.5$      &$81.6$ &$98.0$   &$92.7$   &$89.8$   &{$92.1$}\cr
LSTM PM\cite{luo2018lstm}    			&$98.2$ &$96.5$ &$89.6$ &$86.0$ &$98.7$ &$95.6$  &$90.0$   &{$93.6$}\cr
DKD(ResNet-50)\cite{nie2019dynamic}     &$98.3$ &$96.6$ &$90.4$ &$87.1$ &$ 99.1$ &$ 96.0$  &$92.9$   &{$94.0$}\cr
K-FPN(ResNet-18)\cite{zhang2020key} 	&$94.7$ &$96.3$ &$ 95.2$ &$90.2$ &$96.4$ &$95.5$  &$93.2$   &{$94.5$}\cr
K-FPN(ResNet-50)\cite{zhang2020key} 	&$95.1$ &$96.4$ &$\bf 95.3$ &$91.3$ &$96.3$ &$95.6$  &$92.6$   &{$94.7$}\cr 
MotionAdaptive \cite{fan2021motion}		&$98.2$ &$97.4$ &$ 91.7$ &$85.2$ &$\bf 99.2$ &$\bf 96.7$  &$92.2$   &{$94.7$}\cr 
\hline
\rowcolor{gray!20}\bf FAMI-Pose (Ours)&$\bf 99.3$&$\bf 98.6$&$ 94.5$&$\bf 91.7$&$\bf 99.2$&$91.8$&$\bf 95.4$&$\bf 96.0$\cr
    \hline
    \end{tabular}}
    \caption{Performance comparisons on the \textbf{Sub-JHMDB} dataset. 
    } \label{jhmdb}
\end{table}

\subsection{Experimental Settings}
\textbf{Datasets}\quad
PoseTrack is a large-scale benchmark for human pose estimation and articulated tracking in videos, containing challenging sequences of people in crowded scenarios and performing rapid movement. The \textbf{PoseTrack2017} dataset includes 514 video sequences with a total of $16,219$ pose annotations. These are split (following the official protocol) into 250, 50, and 214 video sequences for training, validation, and testing. The \textbf{PoseTrack2018} dataset contains $1,138$ video sequences (and $153,615$ pose annotations), with 593 for training, 170 for validation, and 375 for testing. Both datasets are annotated with 15 joints, with an additional label for joint visibility. Training videos provide dense pose annotations in the center 30 frames, and validation videos further provide pose annotations every four frames. The \textbf{Sub-JHMDB} dataset contains 316 videos for a total of $11,200$ frames. Annotations are done for 15 joints but only visible joints are annotated. Three different data splits are performed for this dataset, each with a training to testing ratio of $3:1$. Following previous works \cite{luo2018lstm, nie2019dynamic, zhang2020key}, we report the mean accuracy over the three splits. 

\begin{figure*}
\begin{center}
\includegraphics[width=0.98\linewidth]{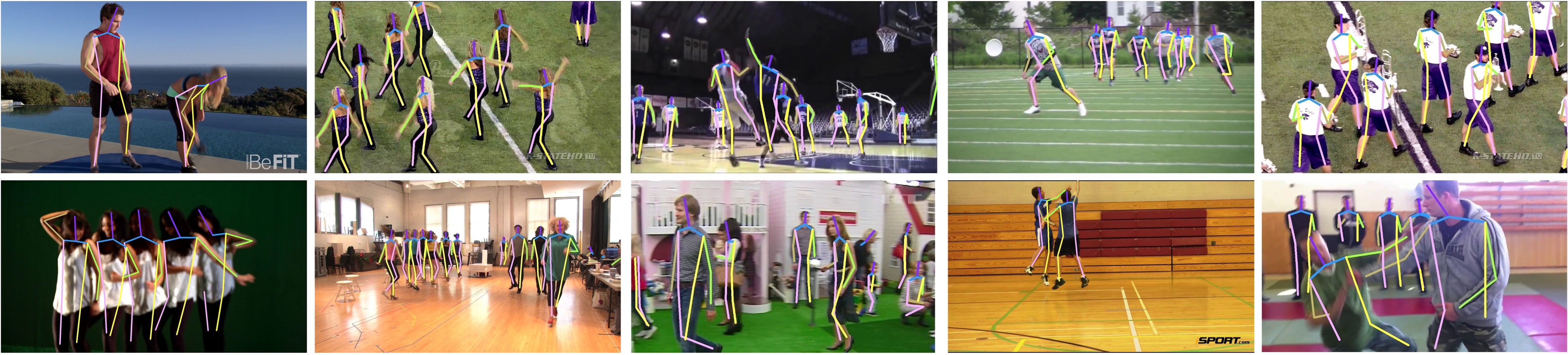}
\end{center}
\caption{Visual results of our FAMI-Pose on benchmark datasets. Challenging scenes such as high-speed motion or pose occlusion are involved.}
\label{fig:results}
\end{figure*}

\textbf{Implementation Details}\quad
Our FAMI-Pose is implemented with PyTorch. The input image size is fixed to $384 \times 288$. We perform data augmentation including random rotation $[-45^{\circ}, 45^{\circ}]$, random scaling $[0.65, 1.35]$, random truncation, and horizontal flipping. The predefined window $\mathcal{N}$ of neighboring frames is set to $\{ -2, -1, 1, 2 \}$, \emph{i.e.}, 2 previous and 2 future frames. We employ the HRNet-W48 model pre-trained on the COCO dataset for feature extraction. Subsequent weight parameters are initialized from a standard Gaussian distribution, while biases are initialized to 0. We employ the Adam optimizer with a base learning rate of $1e-4$ (decays to $1e-5$, $1e-6$, and $1e-7$ at the $8^{th}$, $12^{th}$, and $16^{th}$ epochs, respectively). Training is done with 4 Nvidia Geforce RTX 2080 Ti GPUs and a batch size of 48. All training process is terminated within 20 epochs. To weigh different losses in Eq. \ref{alpha} and Eq. \ref{beta}, we set $\alpha = 1.0$ and $\beta = 0.1$, and have not densely tuned them.

\textbf{Evaluation Metric}\quad
We benchmark our model using the standard human pose estimation protocol \cite{sun2019deep, xiao2018simple}, namely the average precision (\textbf{AP}) metric. We compute the AP for each body joint, and then average over all joints to get the final results (\textbf{mAP}). Note that only visible joints are calculated in performance evaluation.

\subsection{Comparison with State-of-the-art Approaches}
\textbf{Results on the PoseTrack2017 Dataset}\quad
We first evaluate our model on the PoseTrack2017 validation set and test set. A total of $14$ methods are compared, including PoseTracker \cite{girdhar2018detect}, PoseFlow \cite{xiu2018pose}, JointFlow \cite{doering2018joint}, FastPose \cite{zhang2019fastpose}, TML++  \cite{hwang2019pose}, SimpleBaseline (ResNet-50 $and$ ResNet-152), STEmbedding \cite{jin2019multi}, HRNet \cite{sun2019deep}, MDPN \cite{guo2018multi}, Dynamic-GNN \cite{yang2021learning}, PoseWarper \cite{bertasius2019learning}, DCPose \cite{liu2021deep}, and our FAMI-Pose. Their performance on the PoseTrack2017 validation set is reported in Table \ref{17val}. The proposed FAMI-Pose consistently outperforms existing methods, achieving an mAP of $84.8$. Significantly, our FAMI-Pose is able to improve the mAP by $7.5$ points over the widely adopted backbone network HRNet-W48 \cite{sun2019deep}. Our model also achieves a $2.0$ mAP gain over the previous state-of-the-art approach DCPose \cite{liu2021deep}. In particular, we obtain encouraging improvements for the more challenging joints (\emph{i.e.}, wrist, ankle): with an mAP of $80.0$ $(\uparrow 1.6)$ for wrists and an mAP of $77.0$ $(\uparrow 2.8)$ for ankles.
Another interesting observation is that pose estimation approaches that incorporate neighboring frames (such as PoseWarper and DCPose) outperforms methods that use only the single key frame. This suggests the importance of embracing complementary cues from neighboring frames. 

The quantitative comparisons on the PoseTrack2017 test set are reported in Table \ref{17test}. Since the pose annotations are not publicly available, we upload our model predictions to the PoseTrack official evaluation server: \url{https://posetrack.net/leaderboard.php} to obtain results. FAMI-Pose again surpasses previous state-of-the-art, attaining an mAP of $80.9$ ($\uparrow 1.7$), with an mAP of $81.8$, $77.4$, $79.1$, and $73.6$ for the elbow, wrist, knee, and ankle, respectively. As illustrated in in Fig. \ref{fig:results}, the visualized results for scenes with rapid motion or pose occlusions attest to the robustness of our method. 
More visualized results can be found on our project page\footnote{\url{https://github.com/Pose-Group/FAMI-Pose}}.

\begin{table}
  \resizebox{0.48\textwidth}{!}{
  \begin{tabular}{c|ccc|c|c|c}
    \hline
     Method  &Global Transformation &Local Calibration &MI Loss &Wrist &Ankle &Mean\cr
    \hline
    HRNet \cite{sun2019deep} & & & & 73.3 & 68.5 & 77.3\cr
    (a) &\checkmark & & & $78.1$ & $74.3$ & $82.9$\cr
    (b) &\checkmark &\checkmark & & $79.7$ & $76.0$ & $84.0$\cr
    (c) &\checkmark & \checkmark &\checkmark & $\bf80.0$ & $\bf77.0$ & $\bf84.8$\cr
    \hline
    \end{tabular}}
    \caption{Ablation of different components in FAMI-Pose.} \label{abl-com}
\end{table}

\renewcommand\arraystretch{1.3}
\begin{table}
  \resizebox{0.48\textwidth}{!}{
  \begin{tabular}{c|c|c|c|c|c|c|c|c}
    \hline
     Supp. Frame Window $\mathcal{N}$  &Head&Shoulder &Elbow  &Wrist &Hip &Knee &Ankle &{ Mean}\cr
    \hline
     $\mathcal{N}=\{-1\}$   &$88.1$  &$89.2$ &$83.9$ &$78.0$ &$83.5$ &$80.7$  &$73.4$   &{$ 82.8$}\cr
     $\mathcal{N}=\{-1,1\}$   &$89.1$  &$89.5$ &$84.8$ &$79.0$ &$84.2$ &$82.3$  &$74.9$   &{$ 83.9$}\cr
     $\mathcal{N}=\{-2,-1,1\}$   &$89.3$  &$89.8$ &$85.3$ &$79.8$ &$84.2$ &$82.6$  &$76.2$   &{$ 84.5$}\cr
     $\mathcal{N}=\{-2,-1,1,2\}$   &$\bf 89.6$  &$\bf 90.1$ &$\bf 86.3$ &$\bf 80.0$ &$\bf 84.6$ &$\bf 83.4$ &$\bf 77.0$ &$\bf 84.8$\cr
    \hline
    \end{tabular}}
    \caption{Impact of modifying the supporting frame window.} \vspace{-1em} \label{abl-supp}
\end{table}
\textbf{Results on the PoseTrack2018 Dataset}\quad
We further benchmark our model on the PoseTrack2018 dataset. The detailed results on the validation and test sets are tabulated in Table \ref{18val} and Table \ref{18test}, respectively. From these tables, we observe that our FAMI-Pose consistently attains the new state-of-the-art results for all joints. We obtain a $82.2$ mAP on the validation set and a $79.6$ mAP for the test set. 

\textbf{Results on the Sub-JHMDB Dataset}\quad
Results for the Sub-JHMDB dataset are reported in Table \ref{jhmdb}. 
We observe that existing methods have already achieved an impressive accuracy. Specifically, the current state-of-the-art method MotionAdaptive obtains a $94.7$ mAP on this dataset. In contrast, our method is able to achieve a $96.0$ mAP. We also obtain a $99.3$ mAP for the head joint and a $99.2$ mAP for the hip joint. The $1.3$ mAP improvement over the
already impressive state-of-the-art methods might be an evidence to show the effectiveness of the proposed method.


\begin{figure}
\begin{center}
\includegraphics[width=0.98\linewidth]{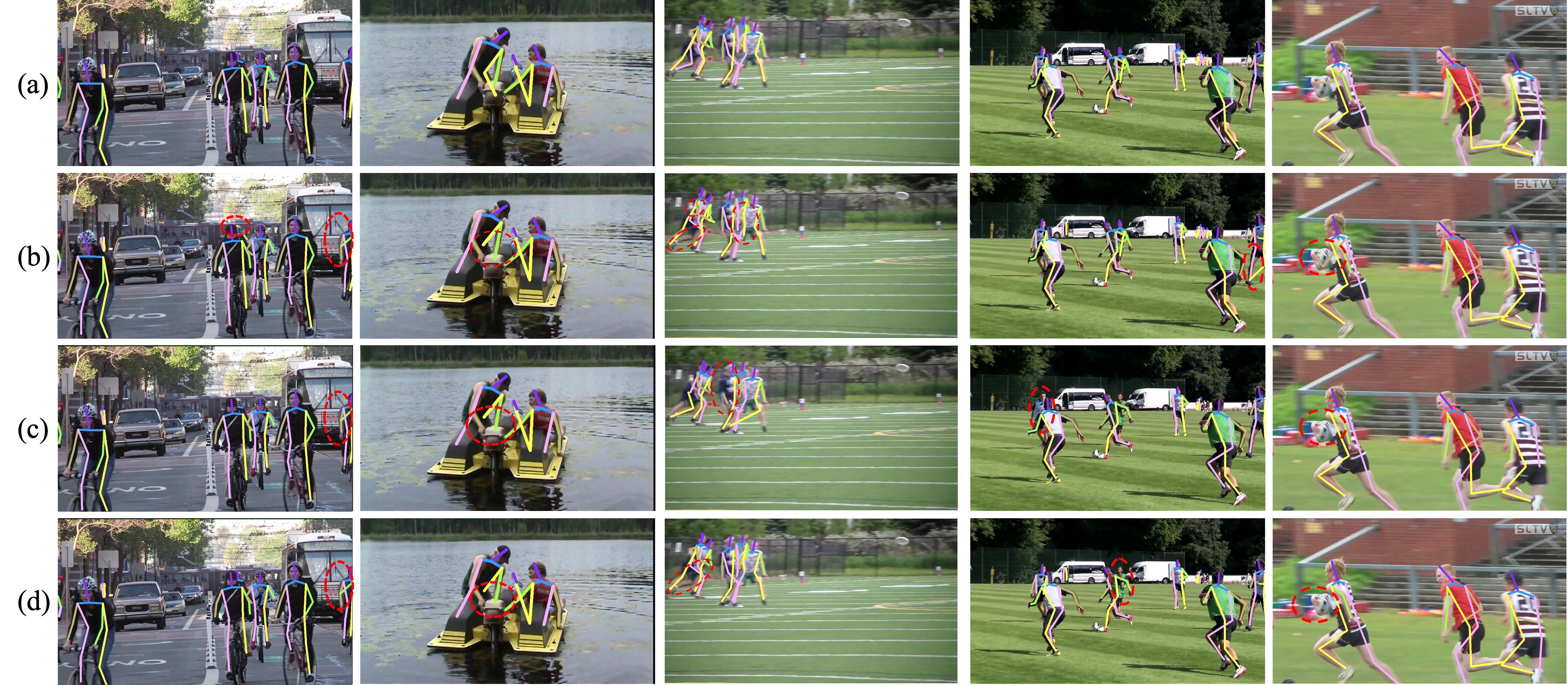}
\end{center}
\vspace{-1em}
\caption{Visual comparisons of the predictions of our FAMI-Pose (a), HRNet-W48 (b), PoseWarper (c), and DCPose (d) on the challenging cases from PoseTrack2017 and PoseTrack2018 datasets. Inaccurate pose estimations are highlighted by the red dotted circles.}\vspace{-1.2em}
\label{fig:contrast}
\end{figure}

\subsection{Ablation Study}
We perform ablation experiments to examine the contribution of feature alignment as well as the influence of each component in our method (\emph{i.e.}, Global Transformation Module, Local Calibration Module, and MI Loss). We also investigate the impact of modifying the predefined window $\mathcal{N}$ of supporting frames. These experiments are conducted on the PoseTrack2017 validation dataset.

\textbf{Feature Alignment}\quad
We empirically evaluate the efficacy of proposed components for facilitating and guiding feature alignment in our FAMI-Pose framework. We report the AP for the wrist and ankle joints as well as the mAP for all joints in Table \ref{abl-com}. 
\textbf{(a)} For the first setting, we remove the local calibration module and MI loss in FAMI-Pose, employing only the global transformation module (GTM) for feature alignment. Remarkably, the coarse feature alignment with the GTM already improves upon the baseline (HRNet-W48 backbone) by a significant margin of $5.6$ mAP and the $82.9$ mAP is in fact on par with the previous state-of-the-art $82.8$ mAP of DCPose \cite{liu2021deep}. This corroborates the effectiveness of our approach in introducing feature alignment to facilitate video-based pose estimation. Feature alignment is noticeably more effective in leveraging temporal information from supporting frames as compared to previous methods which adopt optical flow or motion offset estimations. 
\textbf{(b)} For the next setting, we incorporate the local calibration module (LCM) on top of the global alignment to obtain fine-tuned feature alignment. This fine-tuning improves the mAP by $1.1$ to $84.0$. 
\textbf{(c)} The final setting includes the MI objective and corresponds to our complete FAMI-Pose framework. The improvement of $0.8$ mAP provides empirical evidence that our proposed MI loss is effective as an additional supervision to facilitate the learning of complementary task-specific information in temporal features. 

\textbf{Supporting Frames}\quad
In addition, we investigate the effects of adopting different supporting frame windows  $\mathcal{N}$ for pose estimation. The results in Table \ref{abl-supp} suggest a performance improvement with higher number of supporting frames, whereby the mAP increases from $82.8$ for $\mathcal{N}=\{-1\}$ to $83.9$, $84.5$, $84.8$ at $\mathcal{N}=\{-1,1\}$, $\mathcal{N}=\{-2,-1,1\}$, $\mathcal{N}=\{-2,-1,1,2\}$, respectively. This is in line with our intuitions, \emph{i.e.}, incorporating more supporting frames enables accessing a larger temporal context with more complementary and useful information that are beneficial for improving the pose estimation on the key frame.

\subsection{Comparison of Visual Results}
In addition to the quantitative analysis, we further examine the ability of our model to handle challenging scenarios such as rapid motion or pose occlusions. 
We illustrate in Fig. \ref{fig:contrast} the side-by-side comparisons of a) our FAMI-Pose against state-of-the-art methods, namely b) HRNet-W48 \cite{sun2019deep}, c) PoseWarper \cite{bertasius2019learning}, and d) DCPose \cite{liu2021deep}. It is observed that our approach yields more robust and accurate pose estimates for such challenging scenes. HRNet-W48 is designed for image-based pose estimation and does not incorporate information from supporting frames, resulting in poor performance on degraded video frames. On the other hand, PoseWarper and DCPose implicitly estimate motion cues between frames to improve pose estimation but lack feature alignment and effective supervision on information gain. Through a principled design of the GTM and LCM for progressive feature alignment as well as the MI objective to enhance complementary information mining, FAMI-Pose shows a better ability to handle visual degradation.
\section{Conclusion}
In this paper, we examine the multi-frame human pose estimation task from the perspective of effectively leveraging temporal contexts through feature alignment and complementary information mining. We present a hierarchical coarse-to-fine network to progressively align supporting frame feature with the key frame feature. Theoretically, we further introduce a mutual information objective for effective supervision on intermediate features. Extensive experiments show that our method delivers state-of-the-art results on three benchmark datasets, PoseTrack2017, PoseTrack2018, and Sub-JHMDB.

\section{Acknowledgements}
This paper is supported by the National Natural Science Foundation of China (No. 61902348) and the Key R\&D Program of Zhejiang Province (No. 2021C01104).

{\small
\bibliographystyle{ieee_fullname}
\bibliography{References}
}

\end{document}